\documentclass[sigconf]{aamas} 

\usepackage{balance} 
\usepackage{pgfplots}
\usepackage{siunitx}
\usetikzlibrary{patterns}

\setcopyright{ifaamas}
\acmConference[AAMAS '21]{Proc.\@ of the 20th International Conference on Autonomous Agents and Multiagent Systems (AAMAS 2021)}{May 3--7, 2021}{Online}{U.~Endriss, A.~Now\'{e}, F.~Dignum, A.~Lomuscio (eds.)}
\copyrightyear{2021}
\acmYear{2021}
\acmDOI{}
\acmPrice{}
\acmISBN{}

\acmSubmissionID{114}

\title{Optimising Long-Term Outcomes using Real-World \\Fluent Objectives: An Application to Football}

\author{Ryan Beal}
\affiliation{
  \institution{University of Southampton, UK}}
\email{ryan.beal@soton.ac.uk}

\author{Georgios Chalkiadakis}
\email{gehalk@intelligence.tuc.gr}
\affiliation{%
 \institution{Technical University of Crete, Greece}
}

\author{Timothy J. Norman}
\email{t.j.norman@soton.ac.uk}
\affiliation{%
 \institution{University of Southampton, UK}
}

\author{Sarvapali D. Ramchurn}
\email{sdr1@soton.ac.uk}
\affiliation{%
 \institution{University of Southampton, UK}
}

\begin{abstract}

In this paper, we present a novel approach for optimising long-term tactical and strategic decision-making in football (soccer) by encapsulating events in a league environment across a given time frame. 
We model the teams' objectives for a season and track how these evolve as games unfold to give a \emph{fluent objective} that can aid in decision-making games. We develop Markov chain Monte Carlo and deep learning-based algorithms that make use of the fluent objectives in order to learn from prior games and other games in the environment and increase the teams’ long-term performance. Simulations of our approach using real-world datasets from 760 matches shows that by using optimised tactics with our fluent objective and prior games, we can on average increase teams mean expected finishing distribution in the league by up to 35.6\%.

\end{abstract}

\newcommand{\BibTeX}{\rm B\kern-.05em{\sc i\kern-.025em b}\kern-.08em\TeX}

\begin{document}


\pagestyle{fancy}
\fancyhead{}


\maketitle 


\section{Introduction}

There are many examples in the real-world of agents or teams of agents aiming to optimise their performance over long periods of time. These often involve a series of multi-step games that feed into one another as well as other factors in the wider environment. Examples of this includes security games where agents aim to constantly protect facilities against attackers that are able to change their tactics and decisions \cite{paruchuri2008playing,shieh2012protect,kiekintveld2009computing}, as well as in the stock market where agents aim to continually make optimal decisions to make profits in fluid real-world environments \cite{lux1999scaling,bak1997price,kagan1995risk}. 

In this paper, we focus on the long term optimisation of decision-making in team sports. Specifically in games of Association Football (soccer).\footnote{Referred to as just ``football'' throughout this paper.}  Although the models could be applied in a number of domains, football presents us with an interesting challenge where a team of human agents compete against other teams of agents across long periods and the success of teams is not only judged in individual games but how they perform over a season in a league format (supported with many years of real-world datasets). This means that there are a set of teams whom each season play every other team twice, both home and away. Teams are awarded points based on winning, losing or drawing and at the end of the season teams are awarded prize money and other incentives based on their points gained in comparison to all other teams in a league rankings/standings.\footnote{https://www.express.co.uk/sport/football/1300924/Premier-League-prize-money-2020-how-much-Liverpool-earn.} Past work in this area has focused on optimising performance in individual games \cite{beal2020optimising} or for extracting contribution of individual players \cite{beal2020learning,decroos2020vaep,fernandez2019decomposing}. However, to date, there is no formal model for optimising team performance and tactical decision-making over a longer period of time. 

Against this background, we propose a formal model for optimising the long-term performance of football teams and how they can extract knowledge from other games in the league environment. We introduce the novel notion of a \emph{fluent objective} which is a sequence of ``objective variables", each one corresponding to a particular point in the agent's planning horizon (i.e., a game in the game season). We should also clarify that these variables can take the form of a broader goal (e.g., win the league or do not get relegated). We use Markov chain Monte Carlo simulations to help look ahead into the future and allow us to set realistic achievable objectives which add more context to our tactical decision-making in individual games. We also take inspiration from observational learning \cite{borsa2019observational,bandura2008observational,jang1999ensemble} to help teams extract information from other games that happen in the environment and past games they have played themselves. This is used to identify tactical decisions that boost the chances of gaining positive results against given oppositions. As the season progresses, teams learn more as more games unfold --- we encapsulate this into our modelling. Thus, this paper advances the state of the art in the following ways:

\begin{enumerate}
    \item We propose a mathematical model for optimising the long-term performance of human teams and apply this to the game of football.
    \item Using real-world data from 760 real-world football games from the past two seasons of the English Premier League (EPL), we can set the fluent objective based on accurate league simulations and further improve individual game payoffs by using knowledge from prior games. In particular, we show that we can increase teams finishing position on average by up to 2.9 ranks (out of 20).
    \item By using a fluent objective and prior game knowledge we are able to show an increased probability of improved long-term performance in real-world football teams (by up to 35.6\%).
\end{enumerate}

Our results show that by looking ahead and thinking about long-term goals, teams can add more context to the tactical decisions that are made for individual games and thus are more likely to achieve the long-term objectives that they have set. 

The rest of this paper is structured as follows, in Section 2 we provide a background and in Section 3 we discuss how we model long term performance. In Section 4 and 5 we discuss how we calculate the fluent objective and learn from prior games respectively. We run simulation experiments on our models in Section 6 and discuss these in Section 7. Finally, Section 8 concludes. 

\section{Background}

In this section, we review related literature showing other examples of modelling real-world problems. We also give an overview of why long-term football tactics are important, what is involved and discuss how this is approached for individual games in \cite{beal2020optimising}.

\subsection{Related Work}

Here, we explore the related work to how we can model long-term flowing games such as a sports league as well and giving some background into sports tactics literature.

\subsubsection{Modelling the Real-World}

As far as we are aware, the notion and modelling of \emph{fluent objectives} in this paper, which allows us to optimise long-term performance, is entirely novel. However,  it was inspired by work presented in situations and fluents in first-order logic and situation calculus \cite{lin2008situation}. We see this approach being used to create a model for environmental context in \cite{ranganathan2003infrastructure}. The authors model enables context awareness to help build context-aware applications. Similarly in our model, we aim to gain context of the other teams in the environments to help make decisions based on the future league standings. There are also agents reacting to situations in their environment in \cite{sim2003agents}, where agents react to the ever-changing variables in the stock market. 

In our work, we also aim to learn from prior games and other games that happen in the environment to gain a better understanding into what tactics work against given opponents. This is closely related to the work presented in \cite{borsa2019observational}, where the authors explore the notion of ``observation learning" which is is a type of learning that occurs as a function of observing, retaining and imitating the behaviour of another agent. This is applicable to football as if we observe another team perform well against another opponent then we may want to imitate their tactics to help us to win. Other examples of this type of work are shown in \cite{piot2013learning,russell1998learning,silver2016mastering}. 

\subsubsection{Sports Tactics}

In the sports domain, there are examples of work focused on team tactics and decision-making in football and other team sports \cite{beal2019artificial}. In terms of long-term decision-making though the key example of agents being used to optimise this in sport is shown in \cite{matthews2012competing} which presents a successful model for competing in fantasy football games.\footnote{https://fantasy.premierleague.com/help/rules.} Here, the authors use machine learning to predict the performance of individual players and then use deep-reinforcement learning to optimise decisions on a week-by-week basis and look ahead to maximise their chances of success. By doing so, they rank in the top 1\% of human players. In our work, we can take inspiration from this in the real-world and help human coaches and managers make decisions on human footballers. 

We also see examples of tactical papers for sport in \cite{jordan2009optimizing} exploring different risk strategies for play-calling in American Football. As well as some key football papers to help improve human performance and identify high-performing players and pairs of players are shown in \cite{fernandez2019decomposing,decroos2020vaep,beal2020learning}.

To provide more intuitions around long-term decision-making, in the next subsection we give a background to football tactics and their importance to the game as well as the league structure.

\subsection{Long-Term Football Tactics}\label{subsec:tactics-background}

In football, individual games are incredibly important, but what is often overlooked tactically is the impact that each game has over a longer period of time and on the overall league standings. The final league standings is the final position of all teams in a league based on the points they have gained over an $N$ game season. In a standard football league (e.g., English Premier League or German Bundesliga), across a season each team plays each other twice (once home and once away) a win is worth 3 points, a draw 1 point and a loss no points. There are huge intrinsic and financial gains to be made by finishing higher up the table and there are certain milestones that teams aim for to boost their success such as qualification for European competitions.\footnote{http://eightyfivepoints.blogspot.com/2018/03/show-me-money-how-much-is-each-premier.html.} 

The season is often broken down into given ``game-weeks" where all teams play a game within the week. We can therefore breakdown the season into these game-weeks as incremental steps in a game. In each week our team plays a game and a number of other games also take place. We therefore, want to maximise our own performance in our game and learn from other games for the future when we play those teams (see Figure \ref{fig:flowchart}). 

Therefore, in this paper we aim to model teams tactical decisions based on the overall league environment and use \emph{fluent objectives} to add context to our decisions and prior games knowledge to imitate other successful teams. In the next section, we discuss the model that this paper builds on for optimising tactical decision-making in individual games. 

\subsection{Modelling the Game of Football}\label{subsec:extend}

The modelling presented in this paper extends on the formal model for football that is presented in \cite{beal2020optimising} for optimising the tactics in an individual game. In \cite{beal2020optimising} the authors use a multi-step game to represent the pre-match tactical decisions that are made using a Bayesian game (representing the unknowns of opposition decisions), this then feeds into the in-match decisions made which is modelled as a stochastic game (representing the score-line states in a game). Using these models teams are able to optimise their tactics by up to 16.1\%. 

In this paper, we extend that model by adding context of the wider environment of the league. By using our fluent objective and prior game weightings we can further optimise these tactics to no only improve the chances of a positive result in the individual game but improve the long-term performance of the team in the league standings. 


\section{Modelling Long Term Team Performance}

\begin{figure*}
    \centering
    \includegraphics[scale=0.6]{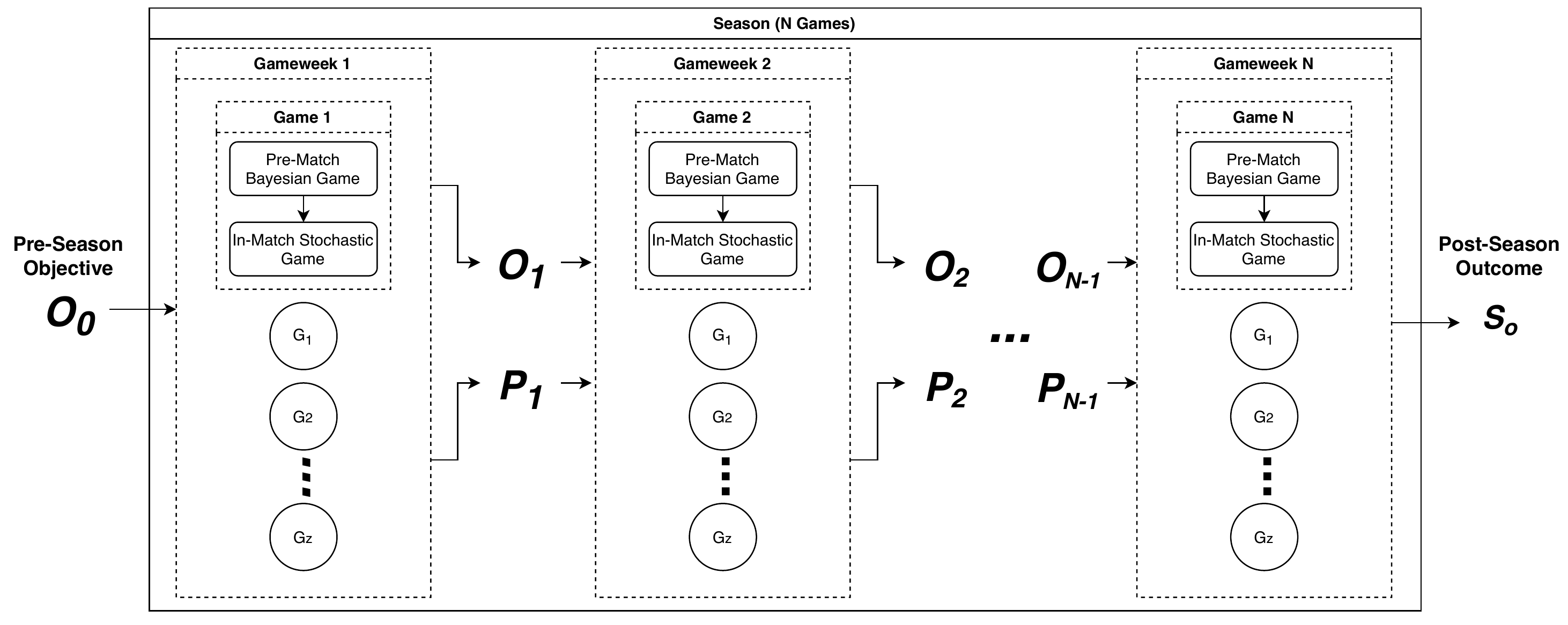}
    \caption{Sequence of Multi-Games Across a Season}
    \label{fig:flowchart}
\end{figure*}


In this section, we discuss how we model the long-term performance of football teams over a season and identify how we can use fluent objectives and learn from games to optimise long-term performance of a team. At the start of a given season or competition, a team will have some aim of how well they want to do and what they want to achieve. In a knockout style cup competition such as the FIFA World Cup or English FA Cup, every team is aiming to win every game as this is the only way to win overall; there are no prizes for second place. Across a full season, however, there are a number of objectives that a team can have that will help maximise their financial gains and reputation of the team. For example, as discussed in Section \ref{subsec:tactics-background}, in the English Premier League there is always only one winner but there are also benefits to finishing in the top 4, top 7 and avoiding finishing in the bottom 3. We therefore, model an entire season in football that could be applied to help optimise teams' long-term performance in any league across the world and at any level. 

\subsection{Sequence of Multi-Games Across a Season}

In Figure \ref{fig:flowchart} we show the structure of our model for an entire season in football. This style of model could also be applied in security games or for emergency response where we aim to optimise the performance of teams of agents in evolving environments with ever-changing objectives \cite{ramchurn2016disaster,shieh2012protect}.

We build on the multi-step (Bayesian into stochastic) games for optimising single game tactics to help teams achieve their objectives in an $N$ game season. There is a sequence of steps that we highlight and show how each one feeds into the next. We also show how a teams' pre-season objective can be fed into the first game which in-turn can use this to aid the tactical decision-making process as well as the parameters we learn while playing each game (e.g., certain tactics that work well against certain teams).

Both the pre-match Bayesian game and the in-match stochastic game can use the objective to help set the risk parameters and select the tactics that will best help the team in the overall environment of the league. This objective then changes as the season progresses and teams aim for different levels of achievement, therefore making this a \emph{fluent objective}; e.g., a team may have had high hopes at the start of the season of winning the league, but if they have a poor start they may have to update their objective to ensure they finish in the top 4. As we show in Figure \ref{fig:flowchart}, the pre-season objective is set as $O_0$, this then changes each game-week as the environment around the team develops, changing to $O_1$ after game-week 1, $O_2$ after game-week 2 and so on until the final in-season objective the week before the final game of the season $N-1$. The final fluent objective, $O_N$, corresponds to the overall end of season outcome ($S_O$), which we can compare to the fluent objective at each game-week to assess the team performance across the season. As discussed in Section \ref{subsec:model-obj}, the $O_x$ and $S_O$ variables might not have distinct values (i.e., maybe $O_0 = O_1$ and so on).

We also consider how we can learn from the games that are played as the season progresses. As we play each game we learn something new, both about what works for our own team and what works against a given opposition. We therefore learn parameters from each game that we can carry forward through each game-week and similarly to the fluent objective we update each week. For example, we may find that when our team uses a given formation against a certain style of opponent we see better results. As we show in Figure \ref{fig:flowchart}, this is encapsulated by a \emph{prior knowledge parameter} $P$, which is updated after each game we play where $P_1$ is after game-week 1, $P_2$ after game-week 2 and so on until the penultimate game-week of the season $O_{N-1}$. We explain the precise form of the $P$ parameter in Section \ref{subsec:prior} below.

Finally, we must consider the other games that are happening each week in the league environment, $\mathcal{G}_N$ is the set of other games in game-week $N$ and $\mathcal{G} = \{G_1, G_2, ..., G_z\} $ where $z$ is the number of other games played in that week. Within each game-week, all other teams also play one another, so that at the end of the season, each team has played every other team twice (once at home and once away). For example, in the EPL there are 20 teams in the league, each team plays the other 19 teams' twice which is 38 games. In the EPL there are a total of 380 games, and so there are 342 that do not involve the team that we are focused on for our optimisation. These games are observable so we can learn from each one, which in turn affects our fluent objective $O$ and what we learn after each game-week $P$. As discussed in Section \ref{subsec:tactics-background}, the outcomes of the other games affect the league table with teams gaining 3 points for a win and 1 point for a draw. We therefore must consider the other teams' performances when setting $O$. We can also observe other games tactically to learn what styles and formations work best against given teams, this is how we can learn $P$ from prior games.

In the following subsections, we go into more detail regarding how we model the fluent objective $O$ and how we can learn from prior games $P$. 

\subsection{Fluent Objectives}\label{subsec:model-obj}

At the start of each season, a team will have some objective for what they are looking to achieve in the next season. These goals are decided based on several factors such as previous season performance and money invested into the team. The goals are usually set by the owners/directors of the team and are based on their subjective opinions of how their team should perform and where they should place in the league against the other teams. The opinions of what the team should achieve then changes over the season which can drive key decisions such as a change in coach/manager for an under-performing team or investing more money into an over-performing team so they achieve a European place which comes with huge financial gains. In other settings, these type of objectives could be the defence of a given target or the rescue of a person. 

Our model for the fluent objective can objectively evaluate how we expect a team to perform over a season and allow teams to change their tactical decision-making based on this. There two different objectives that can be set: a more granular objective of the expected league position and an objective of what could be achieved in terms of broader incentives in the league (e.g., avoiding relegation or qualifying for European competitions). In this paper, we focused on the latter and can define the set of possible objectives as $\mathcal{O} = \{o_1, o_2, ..., o_k\}$ where $k$ is the number of different objectives. An example of the set of objectives --- more accurately, the set of values that an $O_x$ objective variable can take --- in the EPL would be:

\begin{itemize}
    \item \textbf{Winning the League ($o_1$):} Awarded to the team who finishes top of the league.
    \item \textbf{Qualifying for the Champions League ($o_2$):} Awarded to the top 4 teams, so in this case the objective relates to teams finishing 2nd-4th.\footnote{https://www.premierleague.com/european-qualification-explained.}
    \item \textbf{Qualifying for the Europa League ($o_3$):} Another European competition usually awarded to teams who finish between 5th-7th.
    \item \textbf{Top Half Finish ($o_4$):} The financial benefit of finishing higher in the league are huge and therefore teams often aim to finish in the top half of the table (higher than 10th).\footnote{https://www.goal.com/en-gb/news/how-much-money-do-premier-league-2019-20-winners-get/19jbauady17cw1ieojo40yextz.}
    \item \textbf{Avoiding Relegation ($o_5$):} The bottom 3 (18th-20th) teams in the EPL are relegated into the English Football League (EFL) Championship which is the second division of the English football leagues.  
\end{itemize}

To set the objective we can simulate how we expect the season to unfold and create a distribution $\mathcal{D}$ that allows us to use a Maximum a Posteriori (MAP) estimation \cite{gauvain1994maximum} for the probability of the team finishing in each position. This then allows us to calculate a set of probabilities for of a team achieving each objective $\mathcal{P} = \{p(o_1), p(o_2), ..., p(o_k)\}$.  We then set the $O_o$ (for a pre-season objective) as the most likely objective that can be achieved by a team that season.  

This process can then be re-run after each game-week is completed to give the fluent objective $O_1$ to $O_{N-1}$. Our simulation of the league will include the real-results which will get more accurate as the season progresses and we learn more about each team. This will then mean we have a fluent objective that will change as the season progresses. At the end of the season, we can compare $O_0$ to $O_{N-1}$ to the final outcome $S_O$ that the team achieves.

\subsection{Learning From Prior Games}\label{subsec:prior}

As well as the fluent objective, we can also improve the tactical decision-making in our Bayesian and stochastic games by adding prior knowledge $P$ that we learn after each game we play and observe. In more general terms we aim to observe and learn from other successful agents and our own actions. This could also be applicable in swarms of UAVs or imitating other agents trading
in the financial markets settings. 

We can learn a set of weights $\mathcal{W}$ that relate to how effective given style/formation pairs (actions that are made in the multi-step games) that we select in our games are against given oppositions style/formation pairs. These weights are initially set to 1 and are then increased if found to be effective and decrease if found to be ineffective. These can be updated after each game-week and also updated from the other games that we observe. Our $P$ value is defined in Equation \ref{eq:pval}.

\begin{equation}\label{eq:pval}
    P = \left( \begin{array}{ccccc}
w_{11} & w_{12} & w_{13} & \hdots & w_{1j} \\
w_{21} & w_{22} & w_{23} & \hdots & w_{2j} \\
\vdots & \vdots  & \vdots  & \hdots & \vdots	\\	
w_{i1} & w_{i2} & w_{i3} & \hdots & w_{ij}
\end{array} \right)
\end{equation}\\

Where $w \in \mathcal{W}$ and $i$/$j$ is the number of possible style/formation pairs. The columns represent the style/formation pair selected by our team and the rows represent the style/formation selected by the opposition (e.g., $w_{ij}$ is how effective our style/formation pair $i$ is against an opposition using style/formation pair $j$). 

In the following sections, we give more details into how we calculate our fluent objective $O$ and how we can learn the weights that make up $P$. We explore how these are used in the individual football match multi-step game (discussed in Section \ref{subsec:extend}) to further optimise the tactical decision-making process. 

\section{Calculating the Fluent Objective}

In this section, we discuss how we simulate seasons, calculate the fluent objective, and how this can be used to optimise game tactics. 

\subsection{Simulating Season Outcomes}

When we simulate the season outcomes and calculate the distributions of where we expect the team to finish we are interested in predicting all remaining games in the season for both our team and all other teams in the league. To do this we first look at the single-game prediction which is discussed in the next subsection.

\subsubsection{Single-Game Prediction}

To predict the outcomes of single games in the league we use the model that is defined in \cite{beal2020optimising} which is used for calculating the single-game payoffs. The model uses the team's tactical style, potential formation and team strength to give probabilities of a team winning the game. The set of features used are: home team style, away team style, home team formation, away team formation and then team strengths are calculated by using the outputs from the model described in \cite{Dixon_Coles}. The target class is the final result of the game: home team win, away team win or a draw.

Using these features, we train a multi-class classification deep neural network. The neural network is trained using stochastic gradient descent using a categorical cross-entropy loss function (Equation \ref{eq:ccelf}) and a soft-max activation function.
\begin{equation}\label{eq:ccelf}
    -\frac{1}{N}\sum^N_{i=1}\log p_{\textit{model}} [y_i \in O_{y_i}]
\end{equation}

where, $N$ is the number of games that we are using to train the model and $p_{\textit{model}} [y_i \in O_{y_i}]$ is the probability that $y_i$ is in the class $O$. This model takes the given teams, possible playing styles and possible formations to estimate the probability of winning, drawing or losing the game. Using these probabilities we can simulate the outcome of the entire season, this is discussed in the next sub-section. 

\subsubsection{Season Simulation}

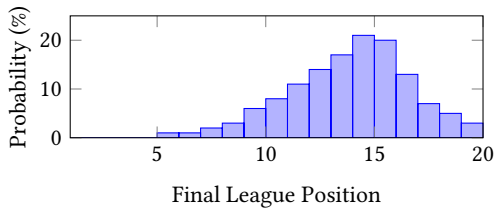
\begin{figure}[h!]
    \centering
    \begin{tikzpicture}
    \begin{axis}[
        ymin=0, ymax=25,
        xmin=1, xmax=20,
        area style,
        xlabel=Final League Position,
        ylabel=Probability (\%),
        width=\columnwidth-40,
        height=\columnwidth-150,
        y label style={at={(axis description cs:0.15,.5)},anchor=south},
        ]
    \addplot+[ybar interval,mark=no] plot coordinates { (1, 0) (2, 0) (3, 0) (4, 0) (5, 1) (6, 1) (7, 2) (8, 3) (9, 6) (10, 8) (11, 11) (12, 14) (13, 17) (14, 21) (15, 20) (16, 13) (17, 7) (18, 5) (19, 3) (20, 2) };
    \end{axis}
    \end{tikzpicture}
    \caption{Example League Outcome Probability Distribution.}
    \label{fig:hist}
\end{figure}

To simulate the remaining games of the season, we use the real-world fixture list to ensure that the ordering of the games is correct. We then find the probability of a home win, away win and draw in each game and use a Markov chain Monte Carlo simulation \cite{vrugt2008accelerating} to simulate all remaining games and total up the points that each team will gain (3 points for a win, 1 for a draw and 0 for a loss). This works well as it emulates the randomness that we see in real-world football games. We repeat this process 100,000 times for each simulation which allows us to derive a distribution for the probability that a team will finish in each place in the league in the final standings. An example of this distribution is shown in Figure \ref{fig:hist}.

\subsection{Setting the Fluent Objective}

Once we have calculated the distributions of possible place outcomes form the MCMC simulation, we use a Maximum a Posteriori (MAP) estimation \cite{gauvain1994maximum} to set the fluent objective. To do this, we can use the posterior distribution to find interval estimates of the final position for the team in the league. We use the position intervals for the objectives discussed in Section \ref{subsec:model-obj} and can find the $o_k \in \mathcal{O}$ that maximises the posterior PDF. This then sets the objective $O_n$ that is used in game-week $n$ and is updated after each game-week.

\subsection{Optimising Tactics using the Fluent Objective}

Once we have set the fluent objective we can now use this when optimising the team tactics in the multi-step game for optimising individual game tactics in that game-week. In the pre-match Bayesian game outlined in \cite{beal2020optimising}, Beal et al. present 3 options that can be used depending on the overall environment. Here we present modified, novel notions of these options, which now employ the fluent objective.

\begin{itemize}
    \item \textbf{Best Response:} Used to maximise the chances of winning a game. This option is selected if a team is currently not on track to achieve their objective for the season and must win games to be able to achieve their goals. 
    \item \textbf{Spiteful:} Used to minimise the chances of the opposition winning the game (and therefore improve your chances of drawing/winning). This option is selected if a team is well ahead of their objective and that by preventing losing the game they are more likely to stay on track for their objective across the season.
    \item \textbf{Expectimax:} This is a mixture of the two above and factors both into account (mathematically defined in \cite{beal2020optimising} where refered to as ``minmax").\footnote{We rename since the approach does not align with the usual meaning of the term ``minimax'' or ``minmax'' in Game Theory.} This is selected if a team is on track for their objective and is aiming to stay that way.  
\end{itemize}

In terms of the in-match stochastic game that is also defined in \cite{beal2020optimising} there are two options that can be selected when making in-match decisions.

\begin{itemize}
    \item \textbf{Aggressive Approach:} This is set if a team is losing/drawing a game and wants to win. It will maximise the chance of a team moving to a more positive state. Therefore, if we know that the objective is to win and gain three points we will select this approach.
    \item \textbf{Reserved Approach:} This is set if a team is winning/drawing and is happy with their current state. It is used to maximise the chances of staying in the current state. Therefore this is used if winning or if a point is a good result in the overall environment in relation to the objective.
\end{itemize}

In the next section, we move on to assess how we can learn from prior games and other games in the environment and how this can be added to our optimising decisions model.




\section{Learning from Previous Games}

In this section, we discuss how we can learn from completed prior games that we play and that other teams in the league play. This allows us to find formation/style combinations that work best against a given formation/style combination that an opposition team may use. To do this we learn a matrix of weights $P$ that corresponds to estimated successes of the formation/style combinations. To estimate each of the weights $w \in P$ we factor in both the games that we have played as well as the games that we have observed. Each weight $w$ corresponds to how effective a given formation/style combination is against a given opposition formation/style. These are computed using Equation \ref{eq:weight} where we look at the games won when using the formation/style ($x$) against the given opposition formation/style ($y$), both in games we have played (first fraction) and in games we have observed (second fraction). 

\begin{equation}\label{eq:weight}
    w_{xy} = \Bigg(\frac{games won}{games played}+\frac{observed games won}{observed games}\Bigg)\div{2}
\end{equation}\\

These weights in $P$ are updated after each game-week so should become more accurate across the season. In game-week 1 all weights can either be set to 1 or be carried over from the previous season. In the next season, we outline how $P$ is used to optimise the pre-game tactics in the Bayesian Game and in-match decisions in the stochastic game.  

\subsection{Optimising Tactics using Prior Games}

Once we have computed the weights that we use in $P$, these can be used when making our pre-match decisions in our Bayesian game. In the optimisation model, a payoff table is computed for each combination of opposition actions to give the probability of the match outcomes based on their selected action of styles $S$ and formations $f$, where $h$ is home win, $d$ is a draw and $a$ is an away win. The payoff for the team is the weighted sum of win and draw probabilities that we store in payoff table made up from the different decision that we can make.

We can then apply the computed weights in $P$ to the payoff table to weigh each payoff depending on how successful these have been in prior games and in observed games. Therefore, we can optimise the tactical decision based on the weighted payoffs in these tables using either the best approach, spiteful or expectimax approaches which are decided based on our fluent objective. This means that if a formation/style combination has never worked in games we have played or observed the payoff will be weighted by 0 and not be selected. The same approach can be applied when changing the formation and style in the in-match stochastic game and each change made can be weighted by the corresponding element in $P$.

In the next section, we perform a number of experiments on our models and assess the performance over a whole given season as well as how the inclusion of $O$ and $P$ each game-week can be used to help teams improve their performance and meet their objectives. 

\section{Empirical Evaluation}

To evaluate our models we use a dataset collected from two seasons (2017/18 and 2018/19) from the English Premier League (EPL).\footnote{All data provided by StatsBomb - www.statsbomb.com.} The dataset breaks down each of the games from the tournament into an event-by-event analysis where each event gives different metrics including event type (e.g., pass, shot, tackle), the pitch coordinates of the event and the event outcome. This type of dataset is industry-leading in football and used by top professional teams.  Thus, this is a rich real-world dataset that allows us to rigorously assess the value of our model.

\subsection{Experiment 1: Learning the Fluent Objective}

Here, we test our fluent objective model in each game-week. Firstly, we evaluate the individual game prediction model that is used to feed the probabilities of outcomes into our season simulation. Secondly, we evaluate our season simulation prediction model using a Markov-chain Monte-Carlo (MCMC) simulation with respect to its accuracy as the season progresses. In Experiment 2, we test our MAP estimator for setting fluent objectives at each game-week. 

To predict the outcome probabilities of individual games we a the deep learning neural network model that calculates pay-offs in the Bayesian game.\footnote{We use a fully-connected feed-forward NN with 3 layers \& a ReLU activation function.} Over the past two EPL seasons the accuracy of the model is 72.99\% with a precision of 69.48\%, recall of 59.5\% and F1 Score of 59.82 \%. This model is used to calculate the probability distribution used in our MCMC model for the entire season. 

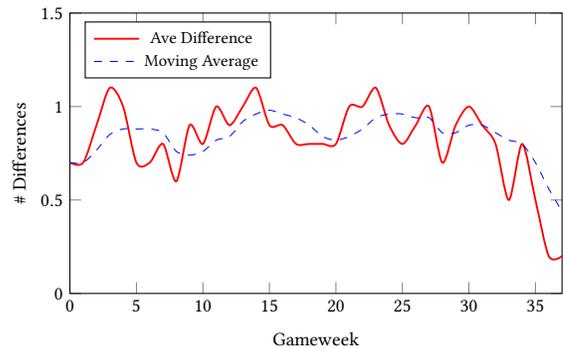
\begin{figure}[h!]
\centering
\begin{tikzpicture}[thick,scale=1, every node/.style={scale=0.8}]
	\begin{axis} [
		xlabel=Gameweek,
		ymin=0,ymax=1.5,
		xmin=0, xmax=37,
		width=\columnwidth-10,
        height=\columnwidth-90,
        legend pos=north west,
        smooth,
        y label style={at={(axis description cs:0.075,.5)},anchor=south},
		ylabel= \# Differences]
		
	\addplot[color=red,line width=0.25mm, solid] coordinates{
		(0,0.7)
        (1,0.7)
        (2,0.9)
        (3,1.1)
        (4,1.0)
        (5,0.7)
        (6,0.7)
        (7,0.8)
        (8,0.6)
        (9,0.9)
        (10,0.8)
        (11,1.0)
        (12,0.9)
        (13,1.0)
        (14,1.1)
        (15,0.9)
        (16,0.9)
        (17,0.8)
        (18,0.8)
        (19,0.8)
        (20,0.8)
        (21,1.0)
        (22,1.0)
        (23,1.1)
        (24,0.9)
        (25,0.8)
        (26,0.9)
        (27,1.0)
        (28,0.7)
        (29,0.9)
        (30,1.0)
        (31,0.9)
        (32,0.8)
        (33,0.5)
        (34,0.8)
        (35,0.5)
        (36,0.2)
        (37,0.2)
	};
	\addplot[color=blue,line width=0.1mm, dashed] coordinates{
		(0,0.7)
        (1,0.7)
        (2,0.7666666666666666)
        (3,0.85)
        (4,0.8800000000000001)
        (5,0.8800000000000001)
        (6,0.8800000000000001)
        (7,0.86)
        (8,0.76)
        (9,0.74)
        (10,0.76)
        (11,0.82)
        (12,0.8400000000000001)
        (13,0.9199999999999999)
        (14,0.9600000000000002)
        (15,0.9800000000000001)
        (16,0.96)
        (17,0.9400000000000001)
        (18,0.9)
        (19,0.8400000000000001)
        (20,0.82)
        (21,0.8400000000000001)
        (22,0.8800000000000001)
        (23,0.9400000000000001)
        (24,0.96)
        (25,0.96)
        (26,0.9400000000000001)
        (27,0.9399999999999998)
        (28,0.86)
        (29,0.8600000000000001)
        (30,0.9)
        (31,0.9)
        (32,0.86)
        (33,0.82)
        (34,0.8)
        (35,0.7)
        (36,0.56)
        (37,0.44000000000000006)
	};
	\addlegendentry{\small Ave Difference}
	\addlegendentry{\small Moving Average}
	\end{axis}
\end{tikzpicture}
\caption{2018/19 EPL Actual League Standings vs MCMC Predictions}
\label{fig:weeks}
\end{figure}

We then run a number of experiments of our MCMC simulation of a season. We predict all remaining games 100,000 times and find the most likely league standings after 38 game-weeks. We can compare this to the final league ranks and compare the differences. In Figure \ref{fig:weeks}, we show an average of all clubs’ absolute difference in their actual finishing position and their predicted finishing position. This is run after each game-week so we have more information about the games that have already been completed. Week 0 is the prediction before any games have been played and week 37 is the final prediction after 37 out of 38 games have been played.  

As shown in Figure \ref{fig:weeks}, we can see how in the first half of the season the league standings remain fairly unpredictable due to the number of different possible combinations that we are attempting to predict --- there are a total of $\num{2.43e+18}$ different combinations of team order that the league could finish in.\footnote{The vast number of possible combination is why we use position differences rather than the overall accuracy of the entire standings after each game-week.} We do see however that as the season unfolds and we have a better idea of team performance the simulation accuracy improves. This is also to be expected as we are simulating fewer games later into the season and we have more evidence from those having taken place in the real world. This shows that we have a suitable method to extract a distribution of where we expect a team to finish and can therefore derive the fluent objective using a MAP estimation to get our objective. This is shown in the next experiment.

\subsection{Experiment 2: Setting the Fluent Objective}

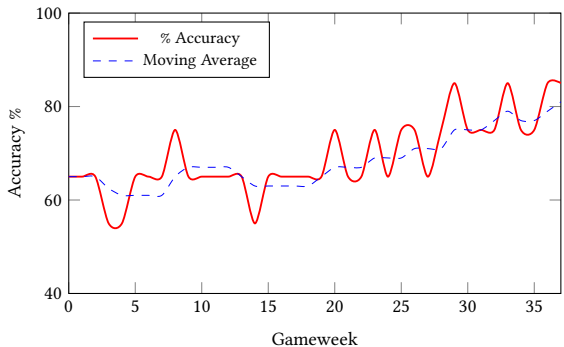
\begin{figure}[h!]
\centering
\begin{tikzpicture}[thick,scale=1, every node/.style={scale=0.8}]
	\begin{axis} [
		xlabel=Gameweek,
		smooth,
		ymin=40,ymax=100,
		xmin=0, xmax=37,
		width=\columnwidth-10,
        height=\columnwidth-90,
        legend pos=north west,
        y label style={at={(axis description cs:0.075,.5)},anchor=south},
		ylabel= Accuracy \%]
		
	\addplot[color=red,line width=0.25mm, solid] coordinates{
		(0,65.0)
        (1,65.0)
        (2,65.0)
        (3,55.00000000000001)
        (4,55.00000000000001)
        (5,65.0)
        (6,65.0)
        (7,65.0)
        (8,75.0)
        (9,65.0)
        (10,65.0)
        (11,65.0)
        (12,65.0)
        (13,65.0)
        (14,55.00000000000001)
        (15,65.0)
        (16,65.0)
        (17,65.0)
        (18,65.0)
        (19,65.0)
        (20,75.0)
        (21,65.0)
        (22,65.0)
        (23,75.0)
        (24,65.0)
        (25,75.0)
        (26,75.0)
        (27,65.0)
        (28,75.0)
        (29,85.0)
        (30,75.0)
        (31,75.0)
        (32,75.0)
        (33,85.0)
        (34,75.0)
        (35,75.0)
        (36,85.0)
        (37,85.0)
	};
	\addplot[color=blue,line width=0.1mm, dashed] coordinates{
		(0,65.0)
        (1,65.0)
        (2,65.0)
        (3,62.5)
        (4,61.0)
        (5,61.0)
        (6,61.0)
        (7,61.0)
        (8,65.0)
        (9,67.0)
        (10,67.0)
        (11,67.0)
        (12,67.0)
        (13,65.0)
        (14,63.0)
        (15,63.0)
        (16,63.0)
        (17,63.0)
        (18,63.0)
        (19,65.0)
        (20,67.0)
        (21,67.0)
        (22,67.0)
        (23,69.0)
        (24,69.0)
        (25,69.0)
        (26,71.0)
        (27,71.0)
        (28,71.0)
        (29,75.0)
        (30,75.0)
        (31,75.0)
        (32,77.0)
        (33,79.0)
        (34,77.0)
        (35,77.0)
        (36,79.0)
        (37,81.0)
	};
	\addlegendentry{\small \% Accuracy}
	\addlegendentry{\small Moving Average}
	\end{axis}
\end{tikzpicture}
\caption{Accuracy of Setting the Fluent Objective (2018/19 EPL Season).}
\label{fig:weeks2}
\end{figure}

To test our MAP estimation, after each game-week simulation we set the fluent objective for all 20 EPL teams. We then assess if their objective was met at that game-week and show the percentage of teams that were successful in meeting their objectives. This is shown in Figure \ref{fig:weeks2} where week 0 is the prediction before any games and week 37 is the final prediction. 


\vspace*{-\baselineskip}

As we can see in Figure \ref{fig:weeks2} the fluent objective accuracy rises as the season progresses and from week 15 onwards we see the accuracy of the fluent objective setting rise more clearly. This shows that we can set realistic to aim for as the season progresses in relation to the actual league outcomes and what was achieved by the teams. One thing to note in this experiment is that not every team in the league can meet their objective as there may be more teams aiming for something than can achieve it (e.g., 3 teams aiming to win the league). Also, 3 teams must always be relegated which the minimum objective is to avoid, meaning that even in the best case only 85\% of teams will achieve their objective. We find that in weeks 36 and 37, we reach the maximum 85\% of teams meeting their objectives.

\subsection{Experiment 3: Learning from Observing Games}

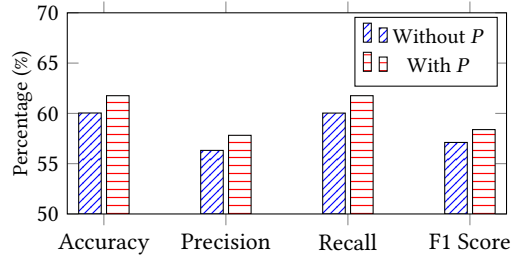
\begin{figure}[h!]
    \centering
    \begin{tikzpicture}
        \centering
        \begin{axis}[
                ybar,
                bar width=0.3cm,
                symbolic x coords={Accuracy,Precision,Recall,F1 Score},
                xtick=data,
                ylabel={\small Percentage (\%)},
                width=\columnwidth-30,
                height=\columnwidth-120,
                ymin=50,ymax=70,
                y label style={at={(axis description cs:0.15,.5)},anchor=south},
            ]
            \addplot[pattern=north east lines, pattern color=blue, 
                every node near coord/.style={inner ysep=5pt}, 
                error bars/.cd, 
                    y dir=both, 
                    y explicit] 
            table[x=interval,y=diff]{\mydata};
            \addplot[pattern=horizontal lines, pattern color=red, 
                every node near coord/.style={inner ysep=5pt}, 
                error bars/.cd, 
                    y dir=both, 
                    y explicit] 
            table[x=interval,y=diff]{\newdata};
        \addlegendentry{\small Without $P$}
    	\addlegendentry{\small With $P$}
        \end{axis}
    \end{tikzpicture}
    \caption{\small Payoff Model Performance Comparison.}
    \label{fig:payoff}
    \end{figure}

To test the impact of the addition of the weights $w$ that we estimate in $P$, we evaluate how the weights are able to boost our ability to predict the outcomes of games based on the tactical decisions and therefore improve our payoff model. To evaluate our $P$ weights, we compare the accuracy of the predictions of the model presented in \cite{Dixon_Coles} both with and without $P$ (this model makes up part of the feature set that is used for calculating the payoffs). We then assess the differences in terms of the models' ability to be able to accurately predict the outcome of the game running the tests over 1046 games. In both cases, the prediction is the highest probability that is given to one of the results (home win, away win and draw). The results from this experiment are shown in Figure \ref{fig:payoff}.\footnote{The precision, recall and F1 score are computed as a weighted average of the ability to predict each outcome using SciKit Learns' multi-class support.}

    \pgfplotstableread[row sep=\\,col sep=&]{
        interval & diff  \\
        Accuracy     & 60.038  \\
        Precision     & 56.32  \\
        Recall    & 60.03 \\
        F1 Score   & 57.11  \\
        }\mydata
        
    \pgfplotstableread[row sep=\\,col sep=&]{
        interval & diff  \\
        Accuracy     & 61.759  \\
        Precision     & 57.82 \\
        Recall    & 61.759 \\
        F1 Score   &  58.388 \\
        }\newdata
    
As we can see in Figure \ref{fig:payoff} by using the weights in $P$ we are able to boost in the accuracy of the model, and therefore the accuracy of our payoffs, achieving a boost of 1.76\%. We also see that there is an increase in the precision, recall and F1 of our model by 1.50\%, 1.72\% and 1.27\% respectively. Even though this represents a fairly small increase to the results of the model in \cite{Dixon_Coles}, it shows that by learning from what tactics have worked (both for your team and others), we can boost our ability to calculate the tactical decision pay-off and therefore our ability to optimise decisions made. Over a large scale of time such as a 38 game-week season, a 1.76\% boost in performance could be the difference between finishing a place higher in the league which can have huge financial gain and help to achieve our set fluent objective.

\subsection{Experiment 4: Optimising Team Long-Term Performance}

\begin{figure}[h!]
    \centering
    \begin{tikzpicture}[thick,scale=1, every node/.style={scale=0.75}]
    \begin{axis}[ 
    xbar, xmin=0,xmax=4,
    xlabel={Average Difference in Final Position},
    bar width=0.25cm,
    symbolic y coords={
        {With $P$ and $O$},
        Without $P$ and $O$},
    ytick=data,
    width=\columnwidth-45,
    height=\columnwidth-155,
    enlarge y limits={abs=0.5cm},
    legend style={at={(0.675,0.05)},anchor=south west}
    ]
    \addplot[pattern=vertical lines, pattern color=red] coordinates {
        (3.735375,{With $P$ and $O$})
        (0.83175,Without $P$ and $O$)};
    \end{axis}
    \end{tikzpicture} 
    \caption{\small Payoffs of Real-World vs. Optimised Decisions}
    \label{fig:boost}
\end{figure}
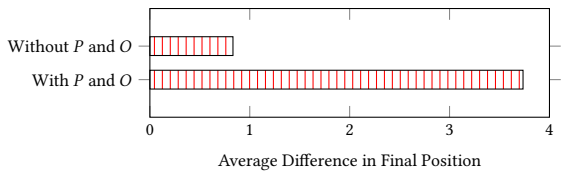

Our final experiment assesses how we incorporate the fluent objective $O$ and weights in $P$ into the tactical decision-making optimisation model presented in \cite{beal2020optimising} and evaluate how this improves team performance to help them meet their objective. To test this we simulate an entire season week by week and apply our model to a single team in the simulation. After each game-week we simulate the remaining games and recalculate $O$ and $P$ as outlined in Figure \ref{fig:flowchart}. We then compare our results using the new model across a simulated season against a simulation where we do not use the $O$ and $P$. We show the results from this when running separate simulations for a set of different teams\footnote{We use the bottom 8 teams in the 2018/19 EPL season to show we can improve their performance.} (the team we use is the only team using the new model in each simulation) in Figure \ref{fig:boost}. We show the average difference in the mean-expected finishing position from the distribution of each team that we run our season simulation for, both using the new model and without.

\begin{figure}[h!]
    \centering
    \begin{tikzpicture}
    \begin{axis}[
        ymin=0, ymax=35,
        xmin=1, xmax=20,
        xlabel=Final League Position,
        ylabel=Probability (\%),
        width=\columnwidth-40,
        height=\columnwidth-120,
        y label style={at={(axis description cs:0.15,.5)},anchor=south},
        smooth,
        legend pos=north west,
        ]
   \addplot[color=red,line width=0.25mm] coordinates{ 
                (1,0.0)
                (2,0.0)
                (3,0.0)
                (4,0.1)
                (5,0.1)
                (6,0.5)
                (7,0.7)
                (8,1.0)
                (9,1.9)
                (10,3.2)
                (11,5.4)
                (12,7.0)
                (13,8.8)
                (14,14.7)
                (15,17.1)
                (16,15.6)
                (17,13.6)
                (18,6.4)
                (19,3.6)
                (20,0.3) }\closedcycle;
     \addplot[color=blue,line width=0.25mm] coordinates{ 
                (1,0.0)
                (2,0.0)
                (3,1.2)
                (4,1.7)
                (5,5.4)
                (6,8.5)
                (7,12.5)
                (8,13.2)
                (9,13.5)
                (10,11.5)
                (11,8.7)
                (12,6.3)
                (13,5.2)
                (14,4.7)
                (15,3.9)
                (16,2.3)
                (17,0.7)
                (18,0.7)
                (19,0.0)
                (20,0.0) }\closedcycle;
    \addplot[red,sharp plot,update limits=false,line width=0.25mm, dashed] 
	coordinates {(14.564, 0) (14.564, 100)};
	\addplot[blue,sharp plot,update limits=false,line width=0.25mm, dashed] 
	coordinates {(9.425, 0) (9.425, 100)};
	\node[text=blue] at (102.5,320) {\small $\mu=9.4$};
	\node[text=red] at (155,320) {\small $\mu=14.6$};
	\addlegendentry{\small Without}
	\addlegendentry{\small With}
    \end{axis}
    \end{tikzpicture}
    \caption{Example League Outcome Probability Distribution for Southampton FC in 2018/19.}
    \label{fig:hist-new}
\end{figure}
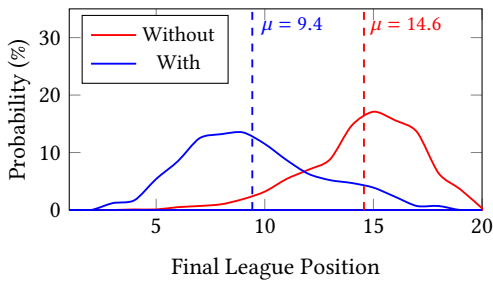

This shows how our model can improve the probability of teams' finishing positions and see that on average there is a 2.90 position improvement when using $O$ and $P$ compared to without for our test set of teams. This is achieved as by using $O$ and $P$ teams can add more context to their decisions, also by selecting the optimal tactics each week in the simulation using the model in \cite{beal2020optimising} we would also expect to see a boost to the performance. Below, we highlight an example of the distribution improvement of the simulation when aiming to optimise the performance of Southampton FC (only team using the optimisation model in the simulation). Figure \ref{fig:hist-new} shows the distribution with $O$ and $P$ applied and not applied.

As we can see from the example shown in Figure \ref{fig:hist-new}, we can use the fluent objectives to help teams boost their probabilities of winning games that matter, and thus boost their expected finishing position, increasing the mean of the expected finishing distribution by up to 35.6\%. We see similar improvements to this across our test set of teams. In the next section, we will further discuss these results, the real-world implications and some further findings.

\section{Discussion}

One interesting finding from further experiments is when we simulate the season with all teams using the model discussed in this paper to select their tactics. When we run this simulation, we find that the results cancels itself out and the final standings are very similar to what we see when we run the simulation without the new fluent objective and prior game weights. We see that there is a boost of under 1 position on average per team when every team uses the model in the same season. This shows that teams can gain a boost in their performance over the season but only if they utilise the game theoretic approaches while all others are not.

Another observation we see in our results is when we compare the increase in the positional distribution using the model between the stronger top-half teams and the teams who are in the lower half of the league and aiming to stay in the division. When using the model for the latter, we observe a substantial boost of up to 35.6\% in long-term performance. This may be due to the algorithm helping teams using the new model gain positive results in the closer games at the bottom of the table when playing teams of similar ability and thus preventing them gaining points by taking all 3 for yourself.  In turn, higher up the league teams often win the games they are expected to against weaker teams so the performance boost is lower. 

It is also worth noting that across the season there are also a number of other variables that can affect team decision-making both tactically and off the pitch. As teams re-assess their objectives in the season, there are decisions off the pitch that can help boost their performing as well as the tactical decision optimisation that helps on it. One example is a change in managers/coaches, this is often a measure taken for an underperforming team and can help boost performance. If a team is doing well and wants to push higher up the table or is struggling and needs new players, then during January teams are able to invest money into new players to improve their team and again improve. These types of decisions could be added into the model to help decision makers at clubs subjectively decide when to invest more money or make changes. 

\section{Conclusions and Future Work}

This paper presents a novel model for the long-term tactical decisions that are made in football and helps teams to optimise their decisions by adding more long-term context. We introduce the concept of a \emph{fluent objective} that allows us to re-evaluate team performance and base decisions based on a wider environment. We find that we can build models that are able to predict the final outcome of the table on a regular basis, and then using a MAP estimation to effectively set the fluent objective each week. We also learn from other games that happen in the overall environment and find that this can boost the performance of pay-off models in our multi-step games. Overall, we find that our model can be used for football teams who are looking to improve their overall expected league position (on average improves teams by 2.90 positions) and, show that the concept of a fluent objective can help to optimise long-term performance in a competitive league setting.

Due to the success we show when using fluent objectives for an application to football in this paper, in future work we intend to test our approach in other domains. For example, they could be used in security games and UAV swarms as the objective also often change over a given time frame. This testing will help to further verify how the modelling of objectives can aid long-term performance. We also aim to further improve our $P$ weights with applications of the observational learning and reinforcement learning as presented in \cite{borsa2019observational}. Finally, the reinforcement learning techniques presented in \cite{silver2016mastering,matthews2012competing} could be used to further optimise team performance. 


\begin{acks}
We would like to thank the reviewers for their comments. This research is supported by the AXA Research Fund and the EPSRC NPIF doctoral training grant number EP/S515590/1.
\end{acks}

\clearpage



\bibliographystyle{ACM-Reference-Format} 
\balance
\bibliography{sample}


\end{document}